\documentclass[runningheads]{llncs}
\usepackage{graphicx}
\usepackage{amsmath,amssymb} % define this before the line numbering.
\usepackage{color}
\usepackage{xcolor}
\usepackage{booktabs} % For formal tables
\usepackage{times}
\usepackage{epsfig}
\usepackage{subfigure}
\usepackage{siunitx}
\usepackage{textcomp}
\usepackage[utf8]{inputenc}
\usepackage[english]{babel}
\usepackage{url}
\usepackage{colortbl}
\usepackage[caption=false]{subfig}

\DeclareMathOperator*{\argmin}{arg\,min}

\begin{document}
\title{Learning to Learn from Web Data through Deep Semantic Embeddings} 
% Replace with your title

\titlerunning{Learning to Learn from Web Data}
% Replace with a meaningful short version of your title
%
\author{Raul Gomez\inst{1,2}\orcidID{0000-0003-4460-3500} \and
Lluis Gomez\inst{2}\orcidID{0000-0003-1408-9803} \and
Jaume Gibert\inst{1}\orcidID{0000-0002-9723-3913} \and
Dimosthenis Karatzas\inst{2}\orcidID{0000-0001-8762-4454}}
%
%Please write out author names in full in the paper, i.e. full given and family names. 
%If any authors have names that can be parsed into FirstName LastName in multiple ways, please include the correct parsing, in a comment to the volume editors:
%\index{Lastnames, Firstnames}
%(Do not uncomment it, because you may introduce extra index items if you do that, we will use scripts for introducing index entries...)
\authorrunning{R. Gomez, L. Gomez, J. Gibert, D. Karatzas}
% Replace with shorter version of the author list. If there are more authors than fits a line, please use A. Author et al.
%

\institute{
Eurecat, Centre Tecnològic de Catalunya, Unitat de Tecnologies Audiovisuals, Barcelona, Spain\\
\email{\{raul.gomez,jaume.gibert\}@eurecat.org}\and
Computer Vision Center, Universitat Autònoma de Barcelona, Barcelona, Spain\\
\email{\{lgomez,dimos\}@cvc.uab.es}} 

\maketitle              % typeset the header of the contribution
\begin{abstract}
In this paper we propose to learn a multimodal image and text embedding from Web and Social Media data, aiming to leverage the semantic knowledge learnt in the text domain and transfer it to a visual model for semantic image retrieval. We demonstrate that the pipeline can learn from images with associated text without supervision and perform a thorough analysis of five different text embeddings in three different benchmarks. We show that the embeddings learnt with Web and Social Media data have competitive performances over supervised methods in the text based image retrieval task, and we clearly outperform state of the art in the MIRFlickr dataset when training in the target data.
Further we demonstrate how semantic multimodal image retrieval can be performed using the learnt embeddings, going beyond classical instance-level retrieval problems.
%Further we evince that multimodal image retrieval works at a semantic level, going beyond classical instance-level retrieval problems where the goal is to retrieve images of an object with no semantics involved. 
Finally, we present a new dataset, InstaCities1M, composed by Instagram images and their associated texts that can be used for fair comparison of image-text embeddings. 

\keywords{self-supervised learning \and webly supervised learning \and text embeddings \and multimodal retrieval  \and multimodal embeddings}
\end{abstract}
\section{Introduction}
\subsection{Why Should We Learn to Learn from Web Data?}
%Deep learning is awesome
Large annotated datasets, powerful hardware and deep learning techniques are allowing to get outstanding machine learning results. Not only in traditional classification problems but also in more challenging tasks such as image captioning or language translation. Deep neural networks allow building pipelines that can learn patterns from any kind of data with impressive results.
%We need tons of data
One of the bottlenecks of training deep neural networks is, though, the availability of properly annotated data, since deep learning techniques are data hungry. Despite the existence of large-scale annotated datasets such as ImageNet \cite{Deng}, COCO \cite{Lin2014} or Places \cite{Zhou2017} and tools for human annotation such as Amazon Mechanical Turk%\footnote{http://www.mturk.com}
, the lack of data limits the application of deep learning to specific problems where it is difficult or economically non-viable to get proper annotations.
%\LGsays{which ones?}. \Rsays{I wrote the kind of problems, sounds weird to write specific ones here}
%Fine-tuning solution

A common strategy to overcome this problem is to first train models in generic datasets as ImageNet and then fine-tune them to other areas using smaller, specific datasets \cite{Yosinski}. But still we depend on the existence of annotated data to train our models. 
%Synthetic data solution
Another strategy to overcome the insufficiency of data is to use computer graphics techniques to generate artificial data inexpensively. However, while synthetic data has proven to be a valuable source of training data for many applications such as pedestrian detection \cite{Marin}, 
%depth and optical flow estimation [Butler2012, Mayer2016], 
image semantic segmentation \cite{Ros2016} 
%human pose estimation [Shotton2013], 
and scene text detection and recognition \cite{Jaderberg2014,Gupta2016}, nowadays it is still not easy to generate realistic complex images for some tasks. 
%Just think about how would you generate synthetic images for different chicken breeds.
%\LGsays{maybe mention here synthetic data as another strategy ... I copy/paste here a paragraph of MINECO in case it is useful: "" An inexpensive source of virtually unlimited data is to synthetically generate images using computer graphics techniques. Synthetic data has proven to be a valuable source of training data for pedestrian detection [Marin2010], depth and optical flow estimation [Butler2012, Mayer2016], image semantic segmentation [Ros2016], human pose estimation [Shotton2013], and scene text detection and recognition [Jaderberg2014, Gupta2016]. }
%Self-supervised solution
%\LGsays{same as before but for Self-supervised approaches ... just mention them in case you want to stress that the problem of learning with less data is something interesting for the community. It's a hot topic!} %\Rsays{Any specific paper I should read and cite?}

%Alternative --> Learn from Web data
An alternative to these strategies is learning from free existing \textit{weakly} annotated multimodal data. Web and Social Media offer an immense amount of images accompanied with other information such as the image title, description or date. This data is noisy and unstructured but it is free and nearly unlimited. Designing algorithms to learn from Web data is an interesting research area as it would disconnect the deep learning evolution from the scaling of human-annotated datasets, given the enormous amount of existing Web and Social Media data. 

\subsection{How to Learn from Web Data?}

\begin{figure}[ht]
\begin{minipage}[c]{0.48\linewidth}\centering
  \includegraphics[width=1\linewidth]{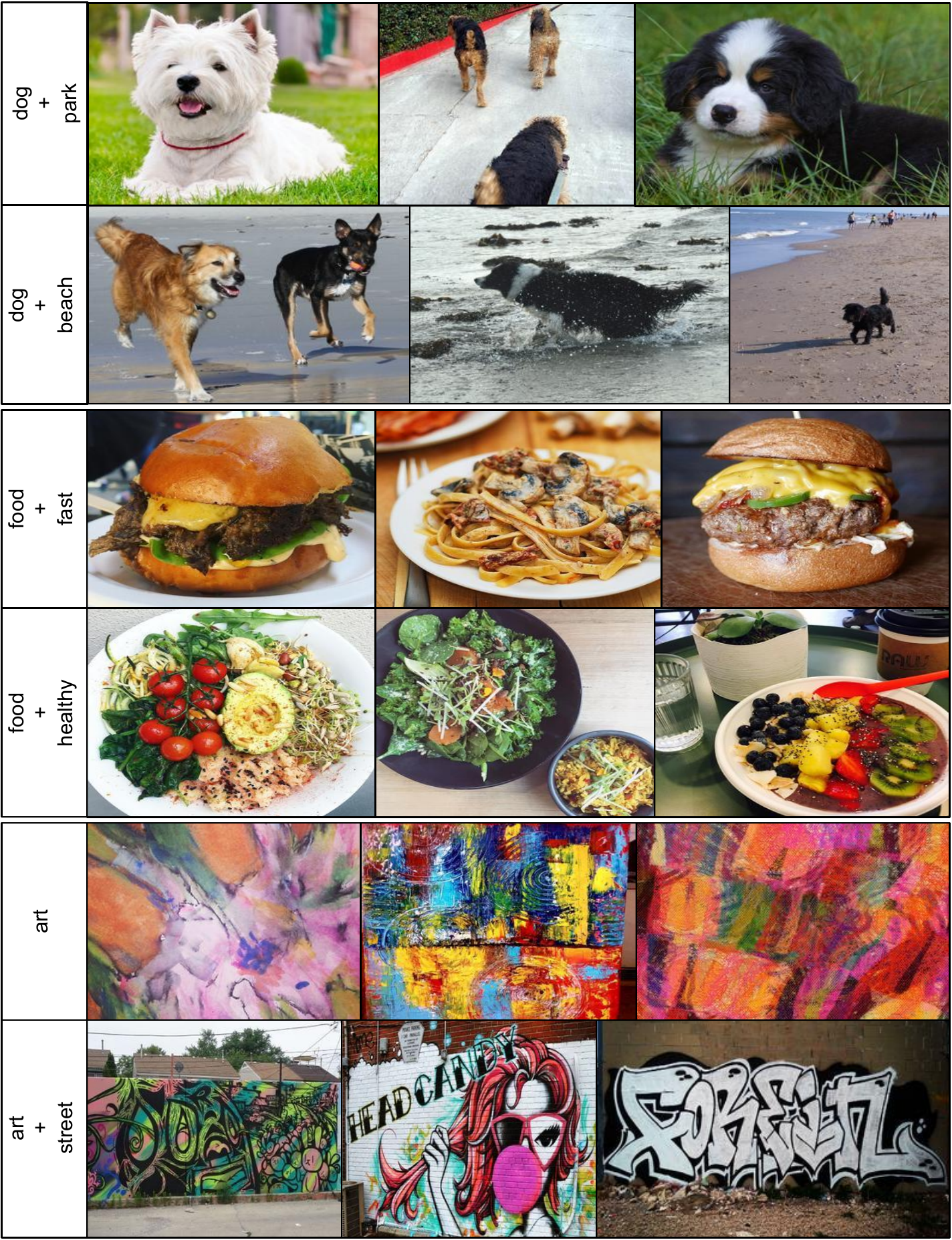}
%   \caption{First retrieved images for complex text queries learning a joint image-text embedding with Word2Vec on InstaCites1M.}
   \caption{Top-ranked results of combined text queries by our semantic image retrieval model. The learnt joint image-text embedding permits to learn a rich semantic manifold even for previously unseen concepts even though they might not be explicitly present in the training set.}
   %The learnt joint image-text embedding permits performing arithmetic operations over individual queries expressed in any of the two modalities.} 
   %Top-ranked results of combined text queries. The learnt joint image-text embedding permits using modifiers in any modality (textual or visual) to bias the results of the original query.
   \label{fig:complex_queries}
\end{minipage}
\hfill
\begin{minipage}[c]{0.48\linewidth}
\centering
  \includegraphics[width=1\linewidth]{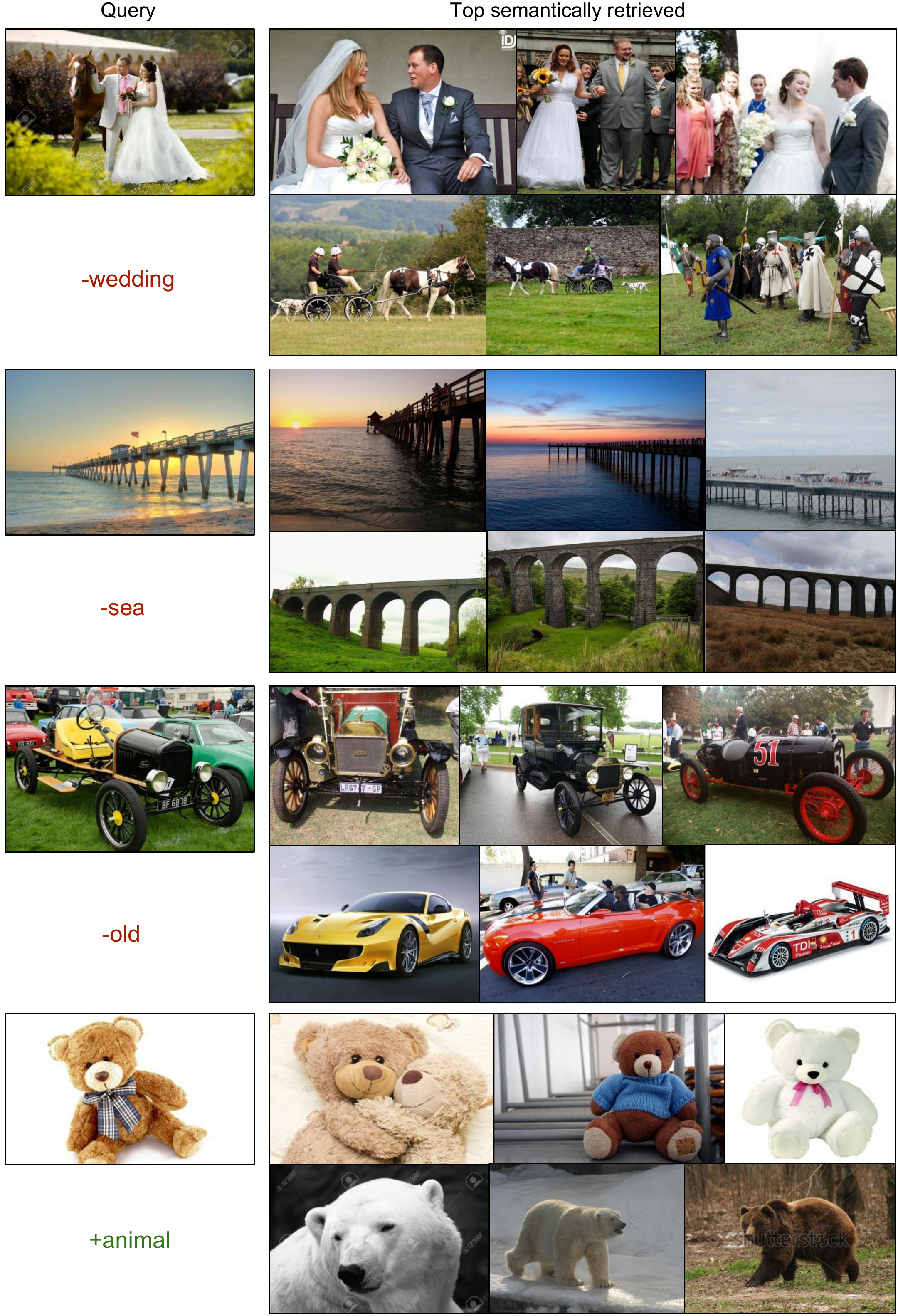}
   \caption{First retrieved images for multimodal queries (concepts are added or removed to bias the results) with Word2Vec on WebVision.}
   \label{fig:image_queries}
   \end{minipage}%
\end{figure}

%Can learn from every word associated to an image, not limited to classes
In some works, such as in the WebVision Challenge \cite{Li2017a}, Web data is used to build a classification dataset: queries are made to search engines using class names and the retrieved images are labeled with the querying class. In such a configuration the learning is limited to some pre-established classes, thus it could not generalize to new classes. While working with image labels is very convenient for training traditional visual models, the semantics in such a discrete space is very limited in comparison with the richness of human language expressiveness when describing an image. 
%Instead we define here a scenario where, by exploiting distributional semantics in a given text corpus, we can learn from every word associated to an image. Thereby we are able to learn which images are associated with, for example, the word ``dolphin", but also with the words ``dolphin in a beach", with the hashtag ``\#healthydinner" or with any piece of text.
Instead we define here a scenario where, by exploiting distributional semantics in a given text corpus, we can learn from every word associated to an image. As illustrated in Figure~\ref{fig:complex_queries}, by leveraging the richer semantics encoded in the learnt embedding space, we can infer previously unseen concepts even though they might not be explicitly present in the training set. %For example, we can learn which images are associated with the phrase "healthy dinner", although in the training set we might only have examples of "healthy exercise" and "romantic dinner".

The noisy and unstructured text associated to Web images provides information about the image content that we can use to learn visual features. A strategy to do that is to embed the multimodal data (images and text) in the same vectorial space. 
%In this work we represent text using five different state of the art methods and we embed it in a space together with image features extracted by a CNN, comparing the performance of the different text embeddings.
In this work we represent text using five different state of the art methods and eventually embed images in the learn semantic space by means of a regression CNN. We compare the performance of the different text space configurations under a text based image retrieval task.
%and show significant improvement over the state of the art in the MIRFlickr dataset. In addition, we introduce InstaCities1M, a new large-scale dataset, and show results in a multi-modal retrieval scenario.
%To evaluate the learned embeddings we set a query by text task: We define some diverse textual queries, embed them in the common text-image space, retrieve the nearest images and check if they are related to the query.

\section{Related Work}
%Multi-modal nets works
Multimodal image and text embeddings have been lately a very active research area. The possibilities of learning together from different kinds of data have motivated this field of study, where both general and applied research has been done. 
DeViSE \cite{Frome2013} proposes a pipeline that, instead of learning to predict ImageNet classes, it learns to infer the Word2Vec \cite{Mikolov2013} representations of their labels. The result is a model that makes semantically relevant predictions even when it makes errors, and generalizes to classes outside of its labeled training set.
Gordo \& Larlus \cite{DianeLarlus2017} use captions associated to images to learn a common embedding space for images and text through which they perform semantic image retrieval. They use a \textit{tf-idf} based BoW representation over the image captions as a semantic similarity measure between images and they train a CNN to minimize a margin loss based on the distances of triplets of query-similar-dissimilar images. Gomez \textit{et al.} \cite{Gomez2017} use LDA \cite{Blei2003} to extract topic probabilities from a bunch of Wikipedia articles and train a CNN to embed its associated images in the same topic space. Wang \textit{et al.} \cite{Wanga} propose a method to learn a joint embedding of images and text for image-to-text and text-to-image retrieval, by training a neural net to embed in the same space Word2Vec \cite{Mikolov2013} text representations and CNN extracted features. 

Other than semantic retrieval, joint image-text embeddings have also been used in more specific applications. Patel \textit{et al.} \cite{Patel} use LDA \cite{Blei2003} to learn a joint image-text embedding and generate contextualized lexicons for images using only visual information. Gordo \textit{et al.} \cite{Gordo} embed word images in a semantic space relying in the graph taxonomy provided by WordNet \cite{PrincetonUniversity2010} to perform text recognition. In a more specific application, Salvador \textit{et al.} \cite{Salvador} propose a joint embedding of food images and its recipes to identify ingredients, using Word2Vec \cite{Mikolov2013} and LSTM representations to encode ingredient names and cooking instructions and a CNN to extract visual features from the associated images.% and learn a joint embedding.
%\LGsays{I miss some references here, I'll tray to list them at some point, ok? I also miss some technical details of the methods you mention. Apart of mentioning the application try to summarize how they work.} 
%\Rsays{I added some technical details, specially about what text embeddings they use}

%Learning from noisy data work
The robustness against noisy data has also been addressed by the community, though usually in an implicit way. Patrini \textit{et al.} \cite{Patrini} address the problem of training a deep neural network with label noise with a loss correction approach and Xiau \textit{et al.} \cite{Xiao} propose a method to train a network with a limited number of clean labels and millions of noisy labels. Fu \textit{et al.} \cite{Fu} propose an image tagging method robust to noisy training data and Xu \textit{et al.} \cite{Xu2017SM} address social image tagging correction and completion. Zhang \textit{et al.} \cite{Zhang2016} show how label noise affects the CNN training process and its generalization error.
%\LGsays{I think the following paper should be mentioned somehow here: "Understanding deep learning requires rethinking generalization"} \Rsays{Interesting paper. I don't know how to fit it in text though}

\subsection{Contributions}
The work presented here brings in a performance comparison between five state of the art text embeddings in multimodal learning, showing results in three different datasets. Furthermore it proves that multimodal learning can be applied to Web and Social Media data achieving competitive results in text-based image retrieval compared to pipelines trained with human annotated data. Finally, a new dataset formed by Instagram images and its associated text is presented: InstaCities1M.
%\LGsays{Here we lack the justification of our work. How it compares to previous work?, why is it important/interesting?}
%In this work there are three main contributions: First, we propose a pipeline to evaluate joint image-text embeddings. Second, we compare the performance of different text embeddings in that pipeline, showing results in three different datasets and proving that the joint embeddings learned solely with Web data have a competitive performance over state of the art supervised methods. Third, we present a dataset formed by Instagram images and its associated text: InstaCities1M.

\section{Multimodal Text-Image Embedding}

%Justification of why we use this pipeline.We don't want to try different CNNs or to add new layers because we want to compare text embeddings
One of the objectives of this work is to serve as a fair comparative of different text embeddings methods when learning from Web and Social Media data. Therefore we design a pipeline to test the different methods under the same conditions, where the text embedding is a module that can be replaced by any text representation.

\begin{figure}[ht]
\centering
{\includegraphics[width=0.65\linewidth]{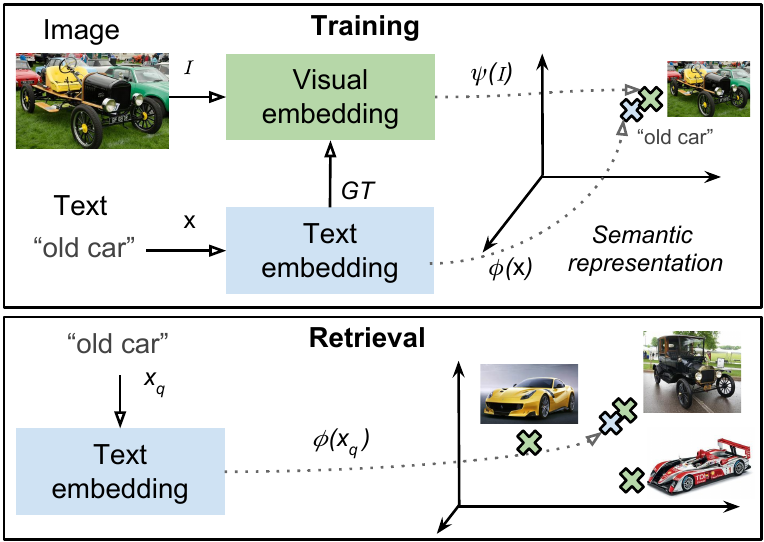}}
\caption{Pipeline of the visual embedding model training and the image retrieval by text.
\label{fig:pipeline}}
\end{figure}

% Pipeline training explanation
The proposed pipeline is as follows: First, we train the text embedding model on a dataset composed by pairs of images and correlated texts $(I, x)$. Second, we use the text embedding model to generate vectorial representations of those texts. Given a text instance $x$, we denote its embedding by $\phi(x) \in \mathbb{R}^d$. Third, we train a CNN to regress those text embeddings directly from the correlated images. Given an image $I$, its representation in the embedding space is denoted by $\psi(I) \in \mathbb{R}^d$. Thereby the CNN learns to embed images in the vectorial space defined by the text embedding model.
The trained CNN model is used to generate visual embeddings for the test set images. Figure \ref{fig:pipeline} shows a diagram of the visual embedding training pipeline and the retrieval procedure.

% Retrieval pipeline explanation
In the image retrieval stage the vectorial representation in the joint text-image space of the querying text is computed using the text embedding model. Image queries can also be handled by using the visual embedding model instead of the text embedding model to generate the query representation. Furthermore, we can generate complex queries combining different query representations applying algebra in the joint text-image space. To retrieve the most semantically similar image $I_R$ to a query $x_q$, we compute the cosine similarity of its vectorial representation $\phi(x_q)$ with the visual embeddings of the test set images $\psi(I_T)$, and retrieve the nearest image in the joint text-image space:
\begin{equation}
\argmin_{I_T \in \text{Test}} \frac{ \langle \phi(x_q) , \psi(I_T) \rangle}{||\phi(x_q)|| \cdot ||\psi(I_T)||}.
\end{equation}

% Why this common embedding rocks
State of the art text embedding methods trained on large text corpus are very good generating representations of text in a vector space where semantically similar concepts fall close to each other. The proposed pipeline leverages the semantic structure of those text embedding spaces training a visual embedding model that generates vectorial representations of images in the same space, mapping semantically similar images close to each other, and also close to texts correlated to the image content. %To do that, the proposed CNN extracts visual patterns from pixels, and then semantic concepts from visual patterns in terms of its vectorial representation. \textcolor{red}{This last sentence in in my mind but I'm not sure I can write it here}
Note that the proposed joint text-image embedding can be extended to other tasks besides image retrieval, such as image annotation, tagging or captioning.

\subsection{Visual Embedding}
A CNN is trained to regress text embeddings from the correlated images minimizing a sigmoid cross-entropy loss. This loss is used to minimize distances between the text and image embeddings. Let $\{(I_n, x_n)\}_{n=1:N}$ be a batch of image-text pairs.  If $\sigma(\cdot)$ is the component-wise sigmoid function, we denote $p_n = \sigma(\phi(x_n))$ and $\hat{p}_n = \sigma(\psi(I_n))$, and let the loss be:
\begin{equation}
L = -\tfrac{1}{N}\sum_{n=1}^{N} [\, p_n \log \hat{p}_n + (1-p_n) \log(1-\hat{p}_n) \, ],
\end{equation}
where the sum's inner expression is averaged over all vector components. 
The GoogleNet architecture \cite{Szegedy} is used, customizing the last layer to regress a vector of  the same dimensionality as the text embedding. %We trained the model only with the last loss head active, since it lead to the same results faster than when using the three default loss heads. 
We train with a Stochastic Gradient Descent optimizer with a learning rate of 0.001, multiplied by 0.1 every 100,000 iterations, and a momentum of 0.9. The batch size is set to 120 and random cropping and mirroring are used as online data augmentation. With these settings the CNN trainings converge around 300K-500K iterations. % and after that they overfit.
We use the Caffe \cite{Jia} framework and initialize with the ImageNet \cite{Deng} trained model to make the training faster. Notice that, despite initializing with a model trained with human-annotated data, this does not denote a dependence on annotated data, since the resulting model can generalize to much more concepts than the ImageNet classes. We trained one model from scratch obtaining similar results, although more training iterations were needed.

%Additionally, the WebVision Challenge \cite{Li2017a} proves that competitive image classification models can be trained solely with Web data. 

%\JGsays{By finetunning the CNN based on ImageNet, we are somehow losing the main point of the work, which is learning exclusively with web data and without supervision at all. The Webvision competition was clearly wanting this. We should at least state that this procedure is to make training faster. More data would avoid having to finetune the CNN and we could train from scratch, but for illustrating the main points of what we want to see (web data can give us resonable results), this is enough. What do you think?} 

%\JGsays{Lately I've been thinking about the learning of text embeddings from scratch using only the training text. If the set of training textual data is not rich enough, it might happen that the embedding is not able to learn some semantics (or just word relations) by itself. I wonder how better could it work if these embeddings are pre-trained using a textual corpus independent from the data in the datasets. In other words, the comparison among different text embeddings might not be thorough if we only depend on the collected data.} 
%\Rsays{Maybe using more textual data improves performance, but I believe that the amount of images with associated text is a harder limitation. The text models seem pretty well trained when checking word distances. We could also compare an initialized text embedding vs one learned from scratch. But that's future work.}

\subsection{Text Embeddings}
%Intro 2 text embeddigns
Text vectorization methods are diverse in terms of architecture and the text structure they are designed to deal with. Some methods are oriented to vectorize individual words and others to vectorize full texts or paragraphs. In this work we consider the top-performing text embeddings and test them in our pipeline to evaluate their performance when learning from Web and Social Media data. Here we explain briefly the main characteristics of each text embedding method used.

%Text embeddings short descriptions
\subsubsection{LDA}\cite{Blei2003}: Latent Dirichlet Allocation learns latent topics from a collection of text documents and maps words to a vector of probabilities of those topics. It can describe a document by assigning topic distributions to them, which in turn have word distributions assigned. An advantage of this method is that it gives interpretable topics.

\subsubsection{Word2Vec}\cite{Mikolov2013}: Using large amounts of unannotated plain text, Word2Vec learns relationships between words automatically using a feed-forward neural network. It builds distributed semantic representations of words using the context of them considering both words before and after the target word.

\subsubsection{FastText}\cite{Bojanowski2016}:  It is an extension of Word2Vec which treats each word as composed of character ngrams, learning representations for ngrams instead of words. The vector for a word is made of the sum of its character n grams, so it can generate embeddings for out of vocabulary words.

\subsubsection{Doc2Vec}\cite{Le2014}: Extends the Word2Vec idea to documents. Instead of learning feature representations for words, it learns them for sentences or documents.

\subsubsection{GloVe}\cite{Pennington}: It is a count-based model. It learns the vectors by essentially doing dimensionality reduction on the co-occurrence counts matrix. Training is performed on aggregated global word-word co-occurrence statistics from a corpus.\\

%\textcolor{red}{Have to learn more about this. See \url{http://clic.cimec.unitn.it/marco/publications/acl2014/baroni-etal-countpredict-acl2014.pdf}}
To the best of our knowledge, this is the first time these text embeddings are trained from scratch on the same corpus and evaluated under the image retrieval by text task.
We used Gensim\footnote{\url{http://radimrehurek.com/gensim}} implementations of LDA, Word2Vec, FastText and Doc2Vec and the GloVe implementation by Maciej Kula\footnote{\url{http://github.com/maciejkula/glove-python}}.
While LDA and Doc2Vec can generate embeddings for documents, Word2Vec, GloVe and FastText only generate word embeddings. To get documents embeddings from these methods, we consider two standard strategies: First, computing the document embedding as the mean embedding of its words. Second, computing a \textit{tf-idf} weighted mean of the words in the document.
%Embeddings dimensionality
For all embeddings a dimensionality of 400 has been used. The value has been selected because is the one used in the Doc2Vec paper \cite{Le2014}, which compares Doc2Vec with other text embedding methods, and it is enough to get optimum performances of Word2Vec, FastText and GloVe, as \cite{Mikolov2013,Bojanowski2016,Pennington} show respectively.
%That value has been selected because has been proved to be one of the optimal ones with huge databases. 
For LDA a dimensionality of 200 has also been considered. 
%\LGsays{yes, we should explain why those values...} \Rsays{400 was selected because I read some publications that said that having a bigger dimensionality did not help if we didn't had also much more data. That can also be seen in the word2vec and GloVe papers. For LDA the 200 value is a random value I used at first and I kept it. Maybe is better to remove it to avoid confusion.}

\section{Experiments}
%Not sure if this should go here
%The trained CNN that embeds images and text in the same space has multiple applications. For instance we could use it in a search by text engine to be able to retrieve not only images which are accompanied with the querying text (as most search engines do now) but also images where the query appears visually. 
%end Not sure if this should go here

\subsection{Benchmarks}
\subsubsection{InstaCities1M}
A dataset formed by Instagram images associated with one of the 10 most populated English speaking cities all over the world (in the images captions one of this city names appears). It contains 100K images for each city, which makes a total of 1M images, split in 800K training images, 50K validation images and 150K test images. The interest of this dataset is that is formed by recent Social Media data. The text associated with the images is the description and the hashtags written by the photo up-loaders, so it is the kind of free available data that would be very interesting to be able to learn from. The InstaCities1M dataset is available on \url{https://gombru.github.io/2018/08/01/InstaCities1M/}. %The InstaCities1M dataset is available for download\footnote{\url{InstaCities1M url}} 

%\Rsays{Can we share those images?}.
%\LGsays{Mmmm ... Looks like it is not legal to redistribute Instagram content: https://www.quora.com/Am-I-allowed-to-share-other-peoples-Instagram-photos-on-my-blog-or-Website/answer/Raj-Jha?srid=dqIP}
%\LGsays{What I've seen in some cases is to provide a list of URLs so people can download the images directly from Instagram servers.}
%\Rsays{Options: URL list (I don't have them, but could collect posts again), Images and text (fair use)}

\subsubsection{WebVision}\cite{Li2017}
It contains more than 2.4 million images crawled from the Flickr Website and Google Images search. The same 1,000 concepts as the ILSVRC 2012 dataset \cite{Deng} are used for querying images. The textual information accompanying those images (caption, user tags and description) is provided. The validation set, which is used as test in this work, contains 50K images.

\subsubsection{MIRFlickr}\cite{Huiskes}
It contains 25,000 images collected from Flickr, annotated using 24 predefined semantic concepts. 14 of those concepts are divided in two categories: 1) strong correlation concepts and 2) weak correlation concepts. The correlation between an image and a concept is strong if the concept appears in the image predominantly. For differentiation, we denote strong correlation concepts by a suffix ``*''. Finally, considering strong and weak concepts separately, we get 38 concepts in total. All images in the dataset are annotated by at least one of those concepts. Additionally, all images have associated tags collected from Flickr. 
Following the experimental protocol in \cite{Lin2015,Xu2017,Li2016,Liu2017} tags that appear less than 20 times are first removed and then instances without tags or annotations are removed.% When using this dataset to test transfer learning results, we take the 5\% of the dataset as the query set and the rest as the retrieval set. The splits made when it is used to train are specified in the experiments section.

\subsection{Retrieval on InstaCities1M and WebVision Datasets}
\subsubsection{Experiment Setup}
To evaluate the learnt joint embeddings, we define a set of textual queries and check visually if the TOP-5 retrieved images contain the querying concept. 
We define 24 different queries. Half of them are single word queries and the other half two word queries. They have been selected to cover a wide area of semantic concepts that are usually present in Web and Social Media data. Both simple and complex queries are divided in four different categories: Urban, weather, food and people. The simple queries are: Car, skyline, bike; sunrise, snow, rain; ice-cream, cake, pizza; woman, man, kid. The complex queries are: Yellow + car, skyline + night, bike + park; sunrise + beach; snow + ski; rain + umbrella; ice-cream + beach, chocolate + cake; pizza + wine; woman + bag, man + boat, kid + dog. For complex queries, only images containing both querying concepts are considered correct.\\ 
%Unnecessary example
%For instance in the query "man + boat" only images containing a man and a boat are considered correct.\\  
%    \LGsays{shall we justify why we do only top-5?} \Rsays{How? I don't think we have to do it since we are comparing methods under the same}

\begin{table}[h]
\noindent
\begin{minipage}[b]{0.48\linewidth}
\caption{Performance on InstaCities1M and WebVision. First column shows the mean P@5 for all the queries, second for the simple queries and third for complex queries.}
%\vspace{0.2cm}
\centering
\resizebox{0.92\linewidth}{!}{
\begin{tabular}{|l|lll|lll|}
\hline
\rowcolor[HTML]{EFEFEF} 
\textbf{Text embedding} & \multicolumn{3}{c|}{\textbf{InstaCities1M}} & \multicolumn{3}{c|}{\textbf{WebVision}}\\ \hline
\rowcolor[HTML]{EFEFEF} 
\textbf{Queries} & \multicolumn{1}{c}{All} & \multicolumn{1}{c}{S} & \multicolumn{1}{c|}{C} & \multicolumn{1}{c}{All} & \multicolumn{1}{c}{S} & \multicolumn{1}{c|}{C} \\ \hline 
\textbf{LDA 200} & 0.40 & 0.73 & 0.07 & 0.11 & 0.18 & 0.03 \\ \cline{1-1}
\textbf{LDA 400} & 0.37 & 0.68 & 0.05 & 0.14 & 0.18 & 0.10 \\ \cline{1-1}
\textbf{Word2Vec mean} & 0.46 & 0.71 & \textbf{0.20} & 0.37 & 0.57 & 0.17 \\ \cline{1-1}
\textbf{Word2Vec tf-idf} & 0.41 & 0.63 & 0.18 & \textbf{0.41} & 0.58 & \textbf{0.23} \\ \cline{1-1}
\textbf{Doc2Vec} & 0.22 & 0.25 & 0.18 & 0.22 & 0.17 & 0.27 \\ \cline{1-1}
\textbf{GloVe} & 0.41 & 0.72 & 0.10 & 0.36 & \textbf{0.60} & 0.12 \\ \cline{1-1}
\textbf{GloVe tf-idf} & \textbf{0.47} & \textbf{0.82} & 0.12 & 0.39 & 0.57 & 0.22 \\ \cline{1-1}
\textbf{FastText tf-idf} & 0.31 & 0.50 & 0.12 & 0.37 & 0.60 & 0.13 \\ \hline 
\end{tabular}}
\vspace{0cm}
\label{tab:results}
\end{minipage}
\hspace{0.3cm}
\begin{minipage}[b]{0.47\linewidth}
\caption{Performance on transfer learning. First column shows the mean P@5 for all the queries, second for the simple queries and third for complex queries.}
%\vspace{0.2cm}
\resizebox{\linewidth}{!}{
\centering
\begin{tabular}{|c|ccc|ccc|}
\hline
\rowcolor[HTML]{EFEFEF} 
\textbf{Text embedding} & \multicolumn{3}{|c|}{\textbf{\begin{tabular}[c]{@{}l@{}}Train: WebVision\\ Test: InstaCities\end{tabular}}} & \multicolumn{3}{c|}{\textbf{\begin{tabular}[c]{@{}l@{}}Train: InstaCities\\ Test: WebVision\end{tabular}}} \\ \hline
\rowcolor[HTML]{EFEFEF} 
\textbf{Queries} & \multicolumn{1}{c}{All} & \multicolumn{1}{c}{S} & \multicolumn{1}{c|}{C} & \multicolumn{1}{c}{All} & \multicolumn{1}{c}{S} & \multicolumn{1}{c|}{C} \\ \hline 
\textbf{LDA 200}  & 0.14 & 0.25 & 0.03 & 0.33 & 0.55 & 0.12 \\ \cline{1-1}
\textbf{LDA 400}  & 0.17 & 0.25 & 0.08 & 0.24 & 0.39 & 0.10 \\ \cline{1-1}
\textbf{Word2Vec mean}  & 0.41 & \textbf{0.63} & 0.18 & 0.33 & 0.52 & 0.15 \\ \cline{1-1}
\textbf{Word2Vec tf-idf}  & \textbf{0.42} & 0.57 & \textbf{0.27} & 0.32 & 0.50 & 0.13 \\ \cline{1-1}
\textbf{Doc2Vec}  & 0.27 & 0.40 & 0.15 & 0.24 & 0.33 & 0.15 \\ \cline{1-1}
\textbf{GloVe}  & 0.36 & 0.58 & 0.15 & 0.29 & 0.53 & 0.05 \\ \cline{1-1}
\textbf{GloVe tf-idf}  & 0.39 & 0.57 & 0.22 & \textbf{0.51} & \textbf{0.75} & \textbf{0.27} \\ \cline{1-1}
\textbf{FastText tf-idf}  & 0.39 & 0.57 & 0.22 & 0.18 & 0.33 & 0.03 \\ \hline
\end{tabular}}
\label{tab:results_transfer}
\end{minipage}
\end{table}

\begin{figure}[h]
\begin{minipage}[c]{0.48\linewidth}
  \includegraphics[width=1\linewidth]{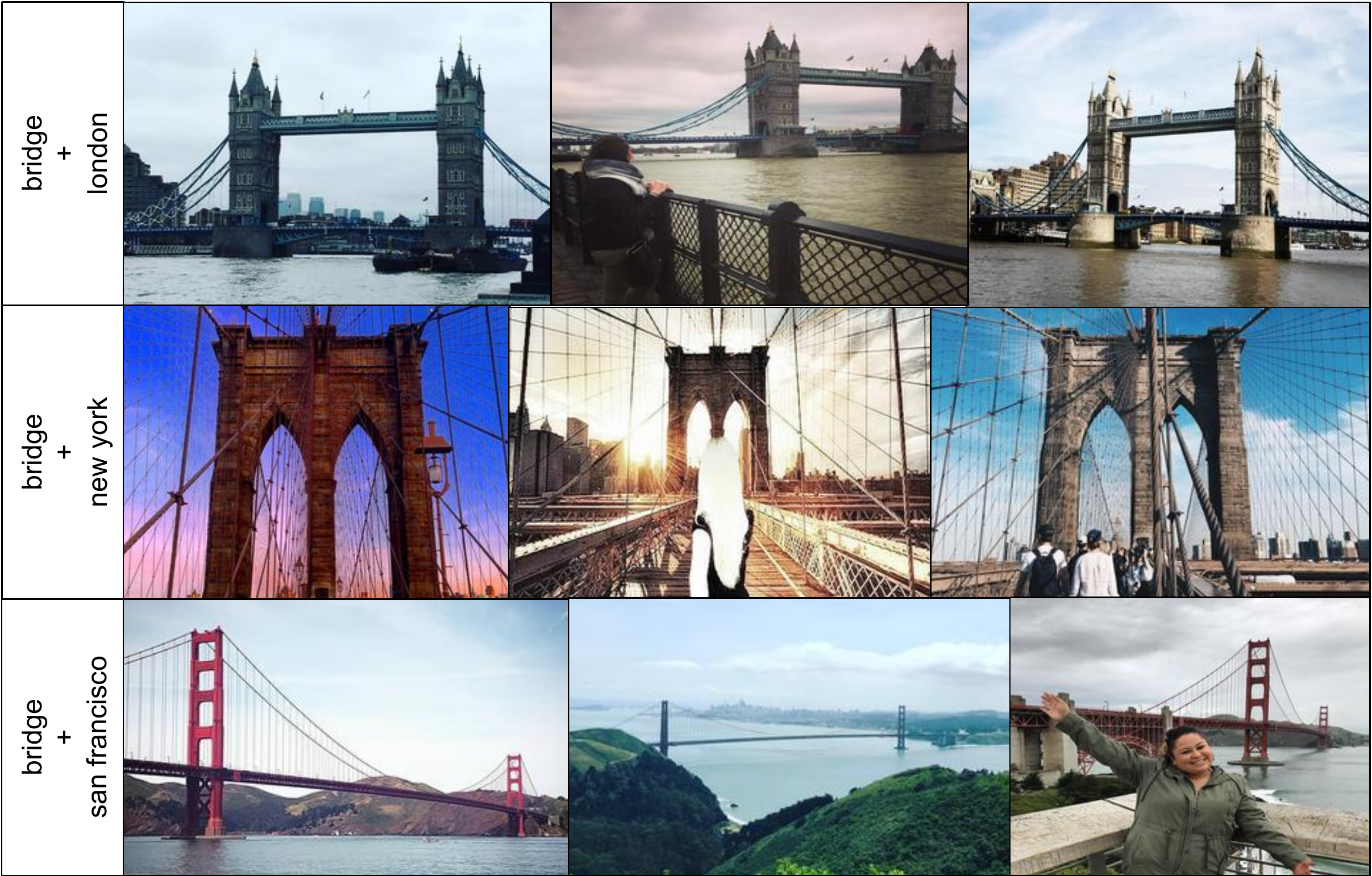}
   \caption{First retrieved images for city related complex queries with Word2Vec on InstaCites1M.}
   \label{fig:bridges}
\end{minipage}
\hfill
\begin{minipage}[c]{0.48\linewidth}
  \includegraphics[width=1\linewidth]{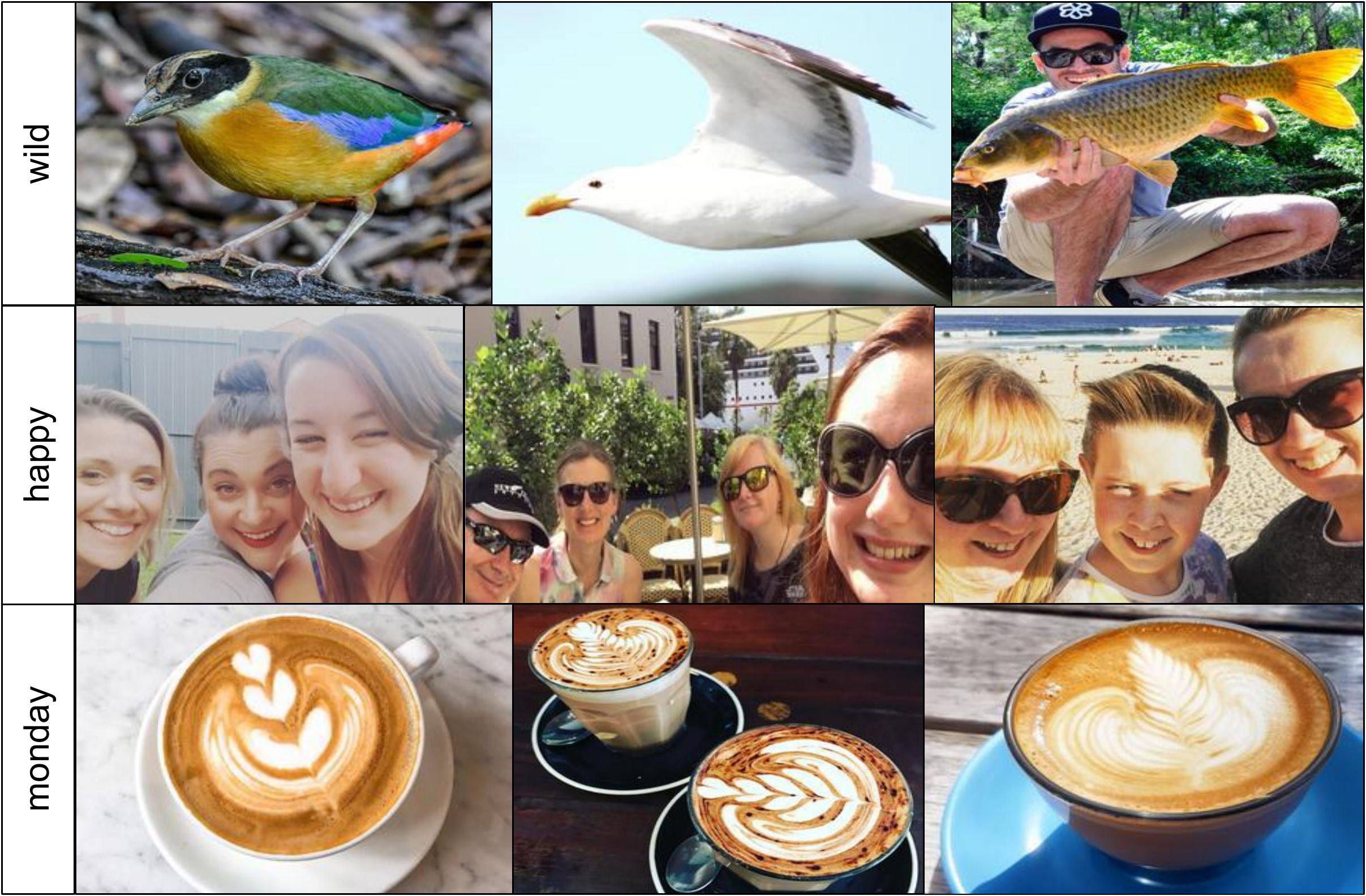}
   \caption{First retrieved images for text non-object queries with Word2Vec on InstaCites1M.}
   \label{fig:abstract_queries}
\end{minipage}%
\end{figure}

%\Rsays{I don't want to use this format for the other figures, because it requires more space and less complex images are shown. But I like it here because it shows images in the middle of two concepts.}

%\begin{figure}
%  \includegraphics[width=\linewidth]{example_4_concepts.pdf}
%   \caption{First retrieved images for simple (corner images) and complex weighted %queries using Word2Vec on InstaCites1M.}
%   \label{fig:example_4_concepts}
%\end{figure}
%\Rsays{Removed since we can't get any example with 3 concepts (only images containing one of the concepts), so this example is the same as the ones using only 2 concepts and more confusing.}

\subsubsection{Results and Conclusions}
%Tables and figures citation
Tables \ref{tab:results} and \ref{tab:results_transfer} show the mean Precision at 5 for InstaCities1M and WebVision datasets and transfer learning between those datasets. To compute transfer learning results, we train the model with one dataset and test with the other. Figures \ref{fig:complex_queries} and \ref{fig:bridges} show the first retrieved images for some complex textual queries. Figure \ref{fig:abstract_queries} shows results for non-object queries, proving that our pipeline works beyond traditional instance-level retrieval.
%Figure \ref{fig:car_train} %and \ref{fig:example_4_concepts} 
%shows how we can weight concepts in the queries to retrieve images between them. 
Figure \ref{fig:image_queries} shows that retrieval also works with multimodal queries combining an image and text.

%\LGsays{I like a lot Figure 5 ... Can we do the same with more words? e.g. a mosaic of 4x4 images with the four images on the corners corresponding to four simple queries ("words")?} \Rsays{I can try to find an example, but will be difficult. First, is difficult to think about concepts that are semantically between other concepts. Then, most of the times images containing both concepts are retrieved (instead of a concept in the middle) if they exist in the database, or images containing only one of the concepts if they don't.} \JGsays{Could we increase the number of images in the weighted mean of the different concepts in the Figures 5 and 6? Even if some images make no sense, it can be interesting to see more steps in the transitions between concepts.} \Rsays{We could but nothing interesting is seen. Usually with complex queries we get either one concept or the other. Only in some cases we achieve getting images containing both concepts or semantically in the middle of them. And when that happens a transition cannot be appreciated. I mean that when we query images between car and train, we get cars, or trains, or people training. We don't get trains that seem cars or anything like that.}

%Complex queries results analysis
For complex queries, where we demand two concepts to appear in the retrieved images, we obtain good results for those queries where the concepts tend to appear together. For instance, we generally retrieve correct images for ``skyline + night'' and for ``bike + park'', but we do not retrieve images for ``dog + kid''. When failing with this complex queries, usually images where only one of the two querying concepts appears are retrieved. Figure \ref{fig:car_train} %and \ref{fig:example_4_concepts} 
shows that in some cases images corresponding to semantic concepts between the two querying concepts are retrieved. That proves that the common embedding space that has been learnt has a semantic structure.  
%Comparing datasets
The performance is generally better in InstaCities1M than in WebVision. The reason is that the queries are closer to the kind of images people tend to post in Instagram than to the ImageNet classes. However, the results on transfer learning show that WebVision is a better dataset to train than InstaCities1M. 
%The results are better when training on WebVision and testing in InstaCities1M than the other way around, and if we compare the transfer learning results with the one we get training and testing with the same dataset, training with WebVision also works better. 
%That's because WebVision has more images than InstaCities1M (2.4M training images vs 800k training images) and shows that the learned models are robust, general and scalable: Having more data, even if it's not specifically related with the task we want to perform, allows learning embedding models that perform better in that task.
%General results analysys. Why LDA fails when training in WebVision.
Results show that all the tested text embeddings methods work quite well for simple queries. Though, LDA fails when is trained in WebVision. That is because LDA learns latent topics with semantic sense from the training data. Every WebVision image is associated to one of the 1,000 ImageNet classes, which influences a lot the topics learning. As a result, the embedding fails when the queries are not related to those classes. 
%Best methods
The top performing methods are GloVe when training with InstaCities1M and Word2Vec when training with WebVision, but the difference between their performance is small. 
%FastText works better on WebVision
FastText achieves a good performance on WebVision but a bad performance on InstaCities1M compared to the other methods. An explanation is that, while Social Media data contains more colloquial vocabulary, WebVision contains domain specific and diverse vocabulary, and since FastText learns representations for character ngrams, is more suitable to learn representations from corpus that are morphologically rich.
%Doc2Vec does not work
Doc2Vec does not work well in any database. That is because it is oriented to deal with larger texts than the ones we find accompanying images in Web and Social Media. %\JGsays{For a fairer comparison with Doc2Vec, it would be nice to find a problem where textual data is of different type. Say the wikipedia articles + the associated images. Since Doc2Vec is thought for embedding 'documents', my intuition is that it will work better than others in situations where the text is richer. It's quite late at this point to introduce another benchmark to the paper, but since we are claiming a comparison among text embedding, we can expect these thoughts from the reviewers.}
%\Rsays{Agree, though I think that it's ok to say that Doc2Vec is not suitable to learn from images with a short caption and hashtags and that's the data we are trying to exploit here}
%Mean vs tf-idf weighted mean
For word embedding methods Word2Vec and GloVe, the results computing the text representation as the mean or as the \textit{tf-idf} weighted mean of the words embeddings are similar.\\

%Failing cases: Word with different meanings, dataset nature, visual features confusion
\subsubsection{Error Analysis}
Remarkable sources of errors are listed and explained in this section.
%Different sources of errors in the retrieval results have been observed:

\paragraph{Visual Features Confusion:} Errors due to the confusion between visually similar objects. For instance retrieving images of a quiche when querying ``pizza''. Those errors could be avoided using more data and a higher dimensional representations, since the problem is the lack of training data to learn visual features that generalize to unseen samples.

\begin{figure}[h]
\begin{minipage}[c]{0.48\linewidth}\centering
  \includegraphics[width=1\linewidth]{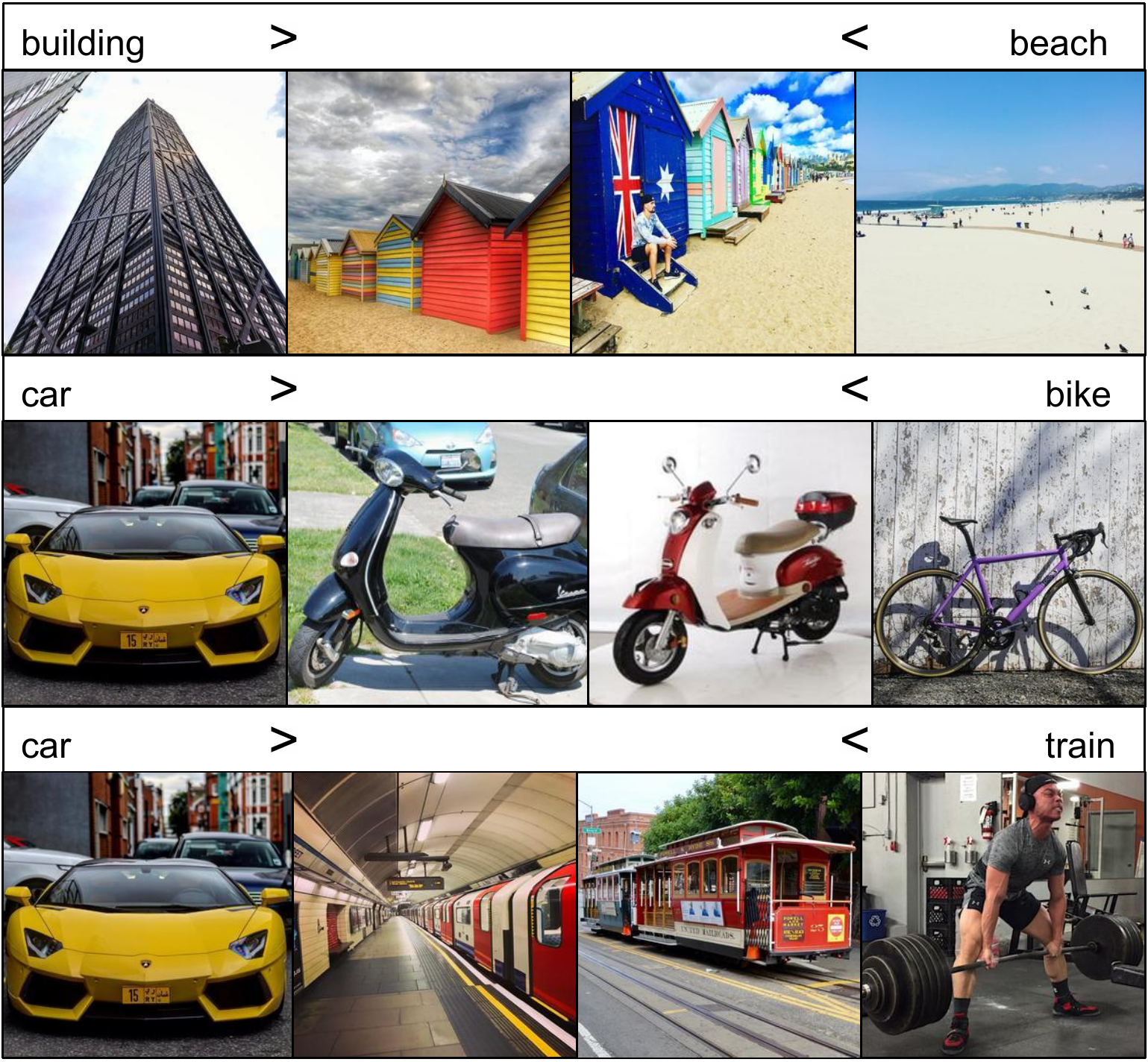}
   \caption{First retrieved images for simple (left and right columns) and complex weighted queries with Word2Vec on InstaCites1M.}
   \label{fig:car_train}
\end{minipage}
\hfill
\begin{minipage}[c]{0.48\linewidth}
\centering
   \includegraphics[width=1\linewidth]{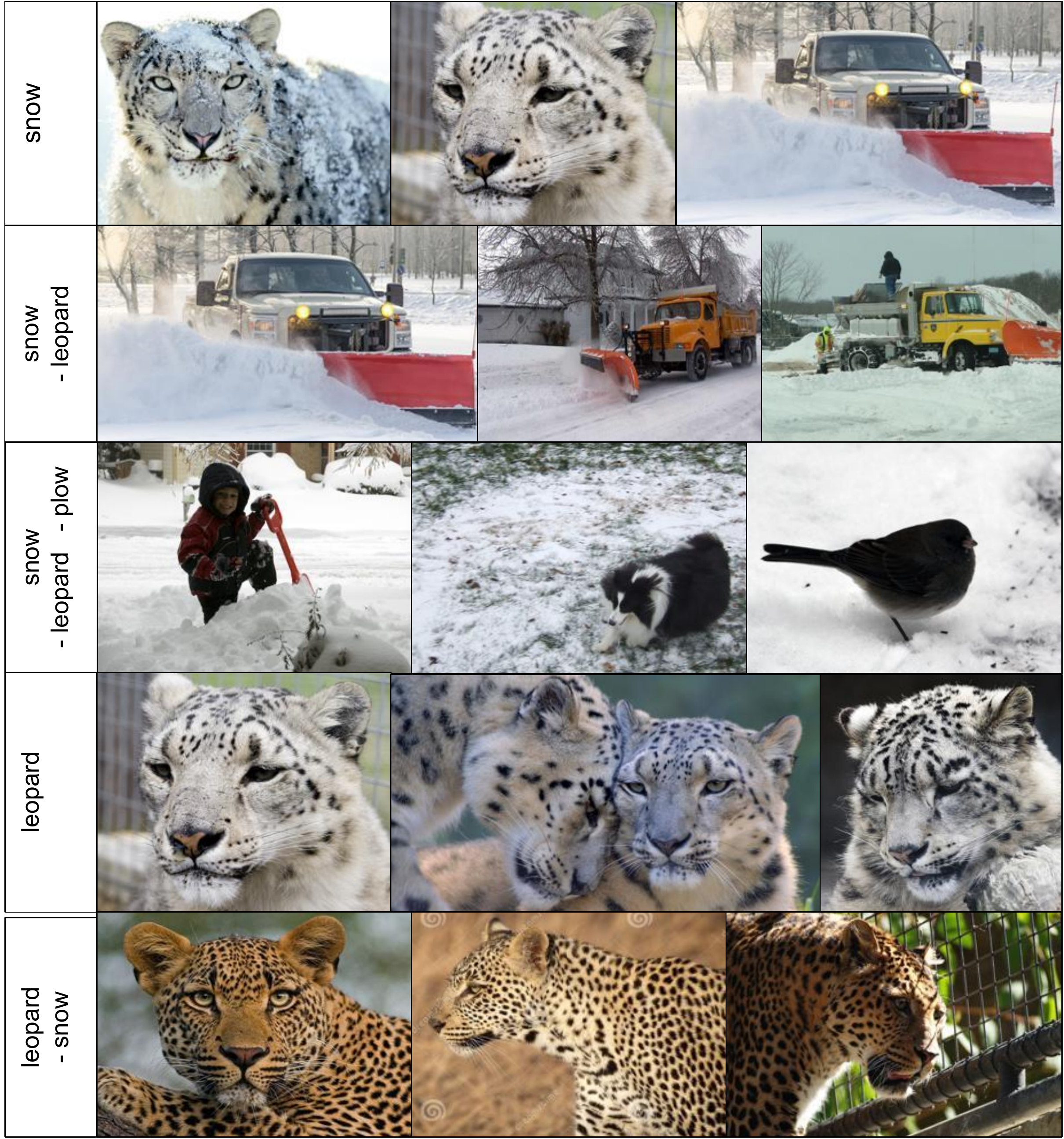}
   \caption{First retrieved images for text queries using Word2Vec on WebVision. Concepts are removed to bias the results.}
   \label{fig:snow_leopard}
\end{minipage}%
\end{figure}

\paragraph{Errors from the Dataset Statistics:} An important source of errors is due to dataset statistics. As an example, the WebVision dataset contains a class which is ``snow leopard'' and it has many images of that concept. The word ``snow'' appears frequently in the images correlated descriptions, so the net learns to embed together the word ``snow'' and the visual features of a ``snow leopard''. There are many more images of ``snow leopard'' than of ``snow'', therefore, when we query ``snow'' we get snow leopard images. Figure \ref{fig:snow_leopard} shows this error and how we can use complex multimodal queries to bias the results.

\paragraph{Words with Different Meanings or Uses:} Words with different meanings or words that people use in different scenarios introduce unexpected behaviors. For instance when we query "woman + bag" in the InstaCities1M dataset we usually retrieve images of pink bags. The reason is that people tend to write "woman" in an image caption when pink stuff appears. 
Those are considered errors in our evaluation, but inferring which images people relate with certain words in Social Media can be a very interesting research.

\subsection{Retrieval in the MIRFlickr Dataset}
%Explain that this is an experiment to be able to compare our results with other pipelines and check that we have learned useful things 
To compare the performance of our pipeline to other image retrieval by text systems we use the MIRFlickr dataset, which is typically used to train and evaluate image retrieval systems. The objective is to prove the quality of the multimodal embeddings learnt solely with Web data comparing them to supervised methods.\\
%Notice that we are using this dataset only to evaluate and not to train our pipeline, which has been solely trained using Web and Social Media data in an unsupervised way. 

%\input{mirflickr.tex}
%&\input{query_classes_new.tex}
\begin{table}
\begin{tabular}{c c}
\resizebox{0.5\linewidth}{!}{

\begin{minipage}[t]{0.48\linewidth}\centering

\caption{MAP on the image by text retrieval task on MIRFlickr as defined in \cite{Xu2017,Liu2017}.}
%\vspace{0.2cm}
\begin{tabular}{|l|l|l|}
\hline
\rowcolor[HTML]{EFEFEF} \textbf{Method}          & \textbf{Train} & \textbf{map}   \\ \hline
\rowcolor[HTML]{FFFFFF} 
\textbf{LDA 200}  & InstaCites1M   & 0.736         \\ \hline
\textbf{LDA 400}  & WebVision      & 0.627          \\ \hline
\textbf{Word2Vec tf-idf} & InstaCites1M   & 0.720          \\ \hline
\textbf{Word2Vec tf-idf} & WebVision      & 0.738          \\ \hline
\textbf{GloVe tf-idf}    & InstaCites1M   & \textbf{0.756} \\ \hline
\textbf{GloVe tf-idf}    & WebVision      & 0.737          \\  \hline
\textbf{FastText tf-idf}    & InstaCities1M      & 0.677          \\  \hline
\textbf{FastText tf-idf}    & WebVision      & 0.734          \\ \hline \hline
\textbf{Word2Vec tf-idf}         & MIRFlickr   & 0.867      \\ \hline
\textbf{GloVe tf-idf}            & MIRFlickr   & \textbf{0.883}      \\ \hline
\textbf{DCH} \cite{Xu2017}             & MIRFlickr  & 0.813 \\ \hline
\textbf{LSRH} \cite{Li2016}            & MIRFlickr   & 0.768          \\ \hline
\textbf{CSDH} \cite{Liu2017}           & MIRFlickr   & 0.764          \\ \hline
\textbf{SePH} \cite{Lin2015}            & MIRFlickr   & 0.735          \\ \hline
\textbf{SCM} \cite{Zhang2014}             & MIRFlickr   & 0.631           \\ \hline
\textbf{CMFH} \cite{Ding2014}            & MIRFlickr  & 0.594           \\ \hline
\textbf{CRH} \cite{Zhen2012}             & MIRFlickr   & 0.581           \\ \hline
\textbf{KSH-CV} \cite{Zhou2014}          & MIRFlickr   & 0.571          \\ \hline
\end{tabular}
\label{tab:mirflickr}

\vspace{1.6cm}
\caption{MAP on the image by text retrieval task on MIRFlickr as defined in \cite{zhang2017}.}
%\vspace{0.2cm}
\begin{tabular}{|l|l|l|}
\hline
\rowcolor[HTML]{EFEFEF} \textbf{Method}          & \textbf{Train} & \textbf{map}   \\ \hline
\rowcolor[HTML]{FFFFFF} 
\textbf{GloVe tf-idf}    & InstaCites1M   & \textbf{0.57} \\ \hline \hline
\textbf{GloVe tf-idf}            & MIRFlickr   & \textbf{0.73}      \\ \hline
\textbf{MML} \cite{zhang2017}             & MIRFlickr  & 0.63 \\ \hline
\textbf{InfR} \cite{zhang2017}            & MIRFlickr   & 0.60          \\ \hline
\textbf{SBOW} \cite{zhang2017}           & MIRFlickr   & 0.59          \\ \hline
\textbf{SLKL} \cite{zhang2017}             & MIRFlickr   & 0.55           \\ \hline
\textbf{MLKL} \cite{zhang2017}            & MIRFlickr  & 0.56           \\ \hline
\end{tabular}
\label{tab:mirflickr2}

\end{minipage}

}

\hspace{0.2cm}

& \begin{minipage}[t]{0.47\linewidth}
\caption{AP scores for 38 semantic concepts and MAP on MIRFlickr. Blue numbers compare our method trained with InstaCities and other methods trained with the target dataset.}
%\vspace{0.2cm}
\centering
\resizebox{0.92\linewidth}{!}{
\begin{tabular}{|l||c|c|c||c|}
\hline
\rowcolor[HTML]{EFEFEF} 

\textbf{Method} 
& \begin{tabular}[c]{@{}c@{}}\textbf{GloVe}\\ \textbf{tf-idf}\end{tabular}   
& \begin{tabular}[c]{@{}c@{}}\textbf{MMSHL}\\ \cite{Wang2017}\end{tabular}  
& \begin{tabular}[c]{@{}c@{}}\textbf{SCM}\\ \cite{Zhang2014}\end{tabular}  
& \begin{tabular}[c]{@{}c@{}}\textbf{GloVe}\\ \textbf{tf-idf}\end{tabular} 
\\ \hline

\rowcolor[HTML]{EFEFEF} 
\textbf{Train} &  \multicolumn{3}{c||}{\textbf{MIRFlickr}}  & \textbf{InstaCities} \\ \hline
\rowcolor[HTML]{FFFFFF} 
\textbf{animals} & \textbf{0.775} & 0.382 & 0.353 & {\color[HTML]{3531FF} 0.707} \\ \hline
\textbf{baby} & \textbf{0.337} & 0.126 & 0.127 & {\color[HTML]{3531FF} 0.264} \\ \hline
\textbf{baby*} & \textbf{0.627} & 0.086 & 0.086 & {\color[HTML]{3531FF} 0.492} \\ \hline
\textbf{bird} & \textbf{0.556} & 0.169 & 0.163 & {\color[HTML]{3531FF} 0.483} \\ \hline
\textbf{bird*} & \textbf{0.603} & 0.178 & 0.163 & {\color[HTML]{3531FF} 0.680} \\ \hline
\textbf{car} & \textbf{0.603} & 0.297 & 0.256 & {\color[HTML]{3531FF} 0.450} \\ \hline
\textbf{car*} & \textbf{0.908} & 0.420 & 0.315 & {\color[HTML]{3531FF} 0.858} \\ \hline
\textbf{female} & \textbf{0.693} & {\color[HTML]{3531FF} 0.537} & 0.514 & 0.481 \\ \hline
\textbf{female*} & \textbf{0.770} & 0.494 & 0.466 & {\color[HTML]{3531FF} 0.527} \\ \hline
\textbf{lake} & \textbf{0.403} & 0.194 & 0.182 & {\color[HTML]{3531FF} 0.230} \\ \hline
\textbf{sea} & \textbf{0.720} & 0.469 & 0.498 & {\color[HTML]{3531FF} 0.565} \\ \hline
\textbf{sea*} & \textbf{0.859} & 0.242 & 0.166 & {\color[HTML]{3531FF} 0.731} \\ \hline
\textbf{tree} & \textbf{0.727} & {\color[HTML]{3531FF} 0.423} & 0.339 & 0.398 \\ \hline
\textbf{tree*} & \textbf{0.894} & 0.423 & 0.339 & {\color[HTML]{3531FF} 0.506} \\ \hline
\textbf{clouds} & \textbf{0.792} & {\color[HTML]{3531FF} 0.739} & 0.698 & 0.613 \\ \hline
\textbf{clouds*} & \textbf{0.884} & 0.658 & 0.598 & {\color[HTML]{3531FF} 0.710} \\ \hline
\textbf{dog} & \textbf{0.800} & 0.195 & 0.167 & {\color[HTML]{3531FF} 0.760} \\ \hline
\textbf{dog*} & \textbf{0.901} & 0.238 & 0.228 & {\color[HTML]{3531FF} 0.865} \\ \hline
\textbf{sky} & \textbf{0.900} & {\color[HTML]{3531FF} 0.817} & 0.797 & 0.809 \\ \hline
\textbf{structures} & \textbf{0.850} & {\color[HTML]{3531FF} 0.741} & 0.708 & 0.703 \\ \hline
\textbf{sunset} & \textbf{0.601} & {\color[HTML]{3531FF} 0.596} & 0.563 & 0.590 \\ \hline
\textbf{transport} & \textbf{0.650} & {\color[HTML]{3531FF} 0.394} & 0.368 & 0.287 \\ \hline
\textbf{water} & \textbf{0.759} & 0.545 & 0.508 & {\color[HTML]{3531FF} 0.555} \\ \hline
\textbf{flower} & \textbf{0.715} & 0.433 & 0.386 & {\color[HTML]{3531FF} 0.645} \\ \hline
\textbf{flower*} & \textbf{0.870} & 0.504 & 0.411 & {\color[HTML]{3531FF} 0.818} \\ \hline
\textbf{food} & \textbf{0.712} & 0.419 & 0.355 & {\color[HTML]{3531FF} 0.683} \\ \hline
\textbf{indoor} & \textbf{0.806} & {\color[HTML]{3531FF} 0.677} & 0.659 & 0.304 \\ \hline
\textbf{plant \_life} & \textbf{0.846} & {\color[HTML]{3531FF} 0.734} & 0.703 & 0.564 \\ \hline
\textbf{portrait} & \textbf{0.825} & {\color[HTML]{3531FF} 0.616} & 0.524 & 0.474 \\ \hline
\textbf{portrait*} & \textbf{0.841} & {\color[HTML]{3531FF} 0.613} & 0.520 & 0.483 \\ \hline
\textbf{river} & \textbf{0.436} & 0.163 & 0.156 & {\color[HTML]{3531FF} 0.304} \\ \hline
\textbf{river*} & \textbf{0.497} & 0.134 & 0.142 & {\color[HTML]{3531FF} 0.326} \\ \hline
\textbf{male} & \textbf{0.666} & {\color[HTML]{3531FF} 0.475} & 0.469 & 0.330 \\ \hline
\textbf{male*} & \textbf{0.743} & {\color[HTML]{3531FF} 0.376} & 0.341 & 0.338 \\ \hline
\textbf{night} & \textbf{0.589} & {\color[HTML]{3531FF} 0.564} & 0.538 & 0.542 \\ \hline
\textbf{night*} & \textbf{0.804} & 0.414 & 0.420 & {\color[HTML]{3531FF} 0.720} \\ \hline
\textbf{people} & \textbf{0.910} & {\color[HTML]{3531FF} 0.738} & 0.715 & 0.640 \\ \hline
\textbf{people*} & \textbf{0.945} & {\color[HTML]{3531FF} 0.677} & 0.648 & 0.658 \\ \hline
\rowcolor[HTML]{EFEFEF} 
\textbf{MAP} & \textbf{0.738} & 0.451 & 0.415 & {\color[HTML]{3531FF} 0.555} \\ \hline
\end{tabular}}
\label{tab:query_classes}
\end{minipage}

\end{tabular}
\end{table}

\subsubsection{Experiment Setup}
%Explain the experiments done
We consider three different experiments: 
1) Using as queries the tags accompanying the query images and computing the MAP of all the queries. Here a retrieved image is considered correct if it shares at least one tag with the query image. For this experiment, the splits used are 5\% queries set and 95\% training and retrieval set, as defined in \cite{Xu2017,Liu2017}.
2) Using as queries the class names. Here a retrieved image is considered correct if it is tagged with the query concept. For this experiment, the splits used are 50\% training and 50\% retrieval set, as defined in \cite{Wang2017}.
3) Same as experiment~1 but using the MIRFlickr train-test split proposed in Zhang et al. \cite{zhang2017}.\\

%Footnotes about dataset splits
%\blfootnote{\textsuperscript{a}The splits used are 5\% queries and 95\% training and retrieval set, as defined in \cite{Xu2017,Liu2017}. We adopted the same splits in order to be able to compare results with them}
%\blfootnote{\textsuperscript{b}The splits used are 5\% queries,  and 95\% retrieval set, from which 5.000 are used also for training, as defined in \cite{Li2016}.}
%\blfootnote{\textsuperscript{c}The splits used are 50\% training and 50\% retrieval set as defined in \cite{Wang2017}.}

\subsubsection{Results and Conclusions}
%Cite results and analyze them
Tables \ref{tab:mirflickr} and \ref{tab:mirflickr2} show the results for the experiments 1 and 3 respectively. We appreciate that our pipeline trained with Web and Social Media data in a multimodal self-supervised fashion achieves competitive results. When trained with the target dataset, our pipeline outperforms the other methods. Table \ref{tab:query_classes} shows results for the experiment 2. Our pipeline with the GloVe \textit{tf-idf} text embedding trained with InstaCites1M outperforms state of the art methods in most of the classes and in MAP. If we train with the target dataset, results are improved significantly. 
%Explain that we are evaluating with this dataset, but we have learned many more things!
Notice that despite being applied here to the classes and tags existing in MIRFlickr, our pipeline is generic and has learnt to produce joint image and text embeddings for many more semantic concepts, as seen in the qualitative examples.\\

\subsection{Comparing the Image and Text Embeddings}

\begin{figure*}[h]
	\centering
  \includegraphics[width=\linewidth]{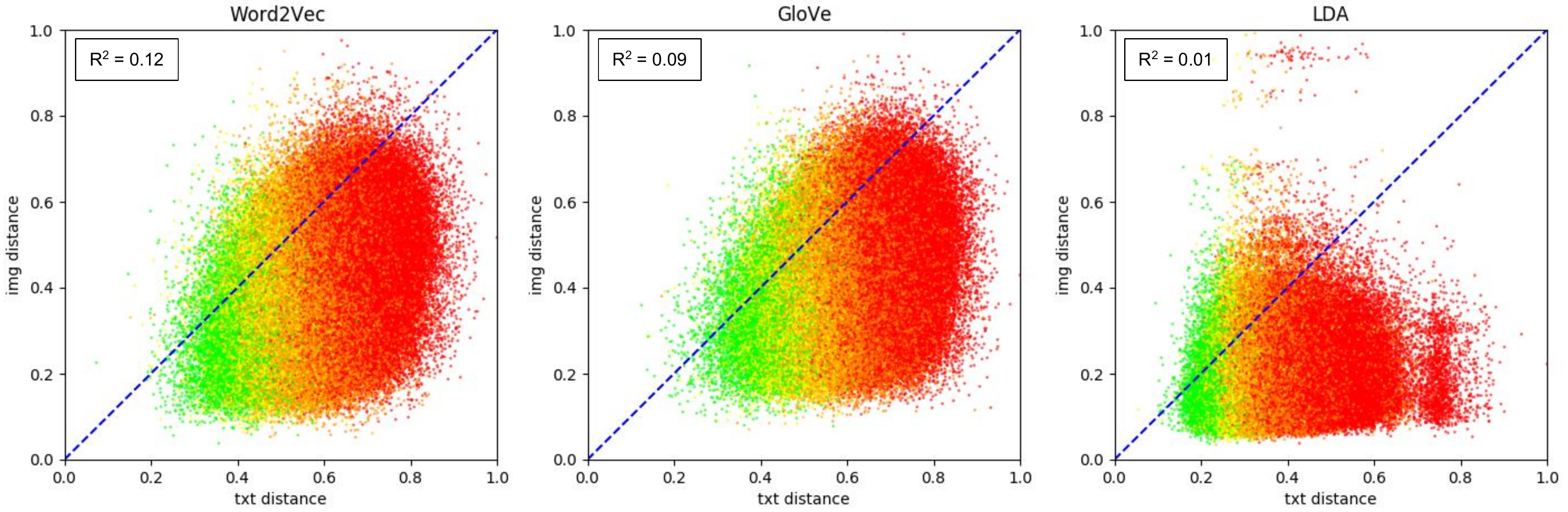}
   \caption{Text embeddings distance (X) vs the images embedding distance (Y) of different random image pairs for LDA, Word2Vec and GloVe embeddings trained with InstaCities1M. Distances have been normalized between [0,1]. Points are red if the  pair does not share any tag, orange if it shares 1, light orange if it shares 2, yellow if it shares 3 and green if it shares more. R\textsuperscript{2} is the coefficient of determination of images and texts distances.}
   \label{fig:embeddings}
\end{figure*}

%This paragraph is unnecessary
%In this work we use Web and Social Media data to learn a joint image and text embedding with a semantic structure. To do that we train a regression CNN to map the images to the text embedding space. 
%end This paragraph is unnecessary

\subsubsection{Experiment Setup}
%Explain how we generate the plots and what we expect to see
To evaluate how the CNN has learnt to map images to the text embedding space and the semantic quality of that space, we perform the following experiment: We build random image pairs from the MIRFlickr dataset and we compute the cosine similarity between 
%both their image embeddings and the embeddings of its associated text. 
both their image and their text embeddings.
In Figure \ref{fig:embeddings} we plot the images embeddings distance vs the text embedding distance of 20,000 random image pairs. 
If the CNN has learnt correctly to map images to the text embedding space, the distances between the embeddings of the images and the texts of a pair should be similar, and points in the plot should fall around the identity line $y=x$. Also, if the learnt space has a semantic structure, both the distance between images embeddings and the distance between texts embeddings should be smaller for those pairs sharing more tags: The plot points' color reflects the number of common tags of the image pair, so pairs sharing more tags should be closer to the axis origin.

%Example to explain the plots
%As an example, an image of a dog with the tags "dog, animal" and an image of a fly with the tags "fly, animal". If the text embedding has been learned correctly, the distance of the projection of those tags in the text embedding space should be bigger than the one between "dog, animal" and "cat, animal", but smaller than the one between other pairs not related at all. If the CNN has correctly learned to embed the images of those animals in the text embedding space, the distance between the dog and the fly image embeddings should be similar than the one between their tags embeddings. So the point given by the pair should fall in the identity line. It should be and nearer to the coordinates origin than the point given by the dog and fly pair, which should also fall in the identity line and nearer to the coordinates origin that another pair that has no relation at all.\\
%Not using MIRFlickr  to train
%Notice that MIRFlickr has been only used to generate the testing pairs, and the only data used to train the embeddings is InstaCities1M.

As an example, take a dog image with the tag "dog", a cat image with the tag "cat" and one of a scarab with the tag "scarab". If the text embedding has been learnt correctly, the distance between the projections of dog and scarab tags in the text embedding space should be bigger than the one between dog and cat tags, but smaller than the one between other pairs not related at all.
If the CNN has correctly learnt to embed the images of those animals in the text embedding space, the distance between the dog and the cat image embeddings should be similar than the one between their tags embeddings (and the same for any pair). So the point given by the pair should fall in the identity line. 
Furthermore, that distance should be nearer to the coordinates origin than the point given by the dog and scarab pair, which should also fall in the identity line and nearer to the coordinates origin that another pair that has no relation at all.\\

\subsubsection{Results and Conclusions}
%Plots analysis: Shape and colors.
%Shape --> CNN maps images well to text embedding space
%Colors --> Space has a semantic sense
%Word2Vec and GloVe
The plots for both the Word2Vec and the GloVe embeddings show a similar shape. The resulting blob is elongated along the $y=x$ direction, which proves that both image and text embeddings tend to provide similar distances for an image pair. The blob is thinner and closer to the identity line when the distances are smaller (so when the image pairs are related), which means that the embeddings can provide a valid distance for semantic concepts that are close enough (dog, cat), but fails inferring distances between weak related concepts (car, skateboard). 
The colors of the points in the plots show that the space learnt has a semantic structure. Points corresponding to pairs having more tags in common are closer to the coordinates origin and have smaller distances between the image and the text embedding. From the colors it can also be deducted that the CNN is good inferring distances for related images pairs: there are just a few images having more than 3 tags in common with image embedding distance bigger than 0.6, while there are many images with bigger distances that do not have tags in common. However, the visual embedding sometimes fails and infers small distances for image pairs that are not related, as those images pairs having no tags in common and an image embedding distance below 0.2.

%LDA
The plot of the LDA embedding shows that the learnt joint embedding is not so good in terms of the CNN images mapping to the text embedding space nor in terms of the space semantic structure. The blob does not follow the identity line direction that much which means that the CNN and the LDA are not inferring similar distances for images and texts of pairs. The points colors show that the CNN is inferring smaller distances for more similar image pairs only when the pairs are very related.

%R2
The coefficient of determination R\textsuperscript{2} measures the proportion of the variance in a dependent variable that is predicted by linear regression and a predictor variable. In this case, it can be interpreted as a measure of how much image distances can be predicted from text distances and, therefore, of how well the visual embedding has learnt to map images to the joint image-text space. It ratifies our plots' visual inspection proving that visual embeddings trained with Word2Vec and GloVe representations have learnt a much more accurate mapping than LDA, and shows that Word2Vec is better in terms of that mapping.

\section{Conclusions}
In this work we learn a joint visual and textual embedding using Web and Social Media data and we benchmark state of the art text embeddings in the image retrieval by text task, concluding that GloVe and Word2Vec are the best ones for this data, having a similar performance and competitive performances over supervised  methods in the image retrieval by text task. 
We show that our models go beyond instance-level image retrieval to semantic retrieval and that can handle multiple concepts queries and also multimodal queries, composed by a visual query and a text modifier to bias the results.
We clearly outperform state of the art in the MIRFlick dataset when training in the target data. 
The code used in the project is available on \url{https://github.com/gombru/LearnFromWebData}.

\section*{Acknowledgments}
This work was supported by the Doctorats Industrials program from the Generalitat de Catalunya, the Spanish project TIN2017-89779-P, the H2020 Marie Skłodowska-Curie actions of the European Union, grant agreement No 712949 (TECNIOspring PLUS), and the Agency for Business Competitiveness of the Government of Catalonia (ACCIO).
\bibliographystyle{splncs04}
\bibliography{mybib}
\end{document}